\documentclass[letterpaper]{article} 
\usepackage{aaai2026}  
\usepackage{times}  
\usepackage{helvet}  
\usepackage{courier}  
\usepackage[hyphens]{url}  
\usepackage{graphicx} 
\usepackage{natbib}  
\usepackage{caption} 
\usepackage{algorithm}

\usepackage{newfloat}
\usepackage{listings}
\DeclareCaptionStyle{ruled}{labelfont=normalfont,labelsep=colon,strut=off} 
\floatstyle{ruled}
\newfloat{listing}{tb}{lst}{}
\floatname{listing}{Listing}
\usepackage{algorithm}
\usepackage[noend]{algpseudocode}
\usepackage[switch]{lineno}
\usepackage{subcaption}
\usepackage{amsthm} 
\usepackage{placeins}
\usepackage{xspace}
\usepackage{xcolor}
\usepackage{amsmath}
\usepackage{amssymb}

\newcommand{\NP}{{\small \ensuremath{\mathsf{NP}}\xspace}}

\usepackage{newfloat}
\usepackage{listings}

\title{Multi-Objective Search: Algorithms, Applications, and Emerging Directions}
\author{
    Oren Salzman\textsuperscript{\rm 1},
    Carlos Hernández Ulloa\textsuperscript{\rm 2},
    Ariel Felner\textsuperscript{\rm 3},
    Sven Koenig\textsuperscript{\rm 4}   
}
\affiliations{
    \textsuperscript{\rm 1}Technion - Israel Institute of Technology\\
    \textsuperscript{\rm 2}Universidad San Sebastian\\
    \textsuperscript{\rm 3}Ben-Gurion University\\
    \textsuperscript{\rm 4}University of California, Irvine\\
    osalzman@cs.technion.ac.il,
    carlos.hernandez@uss.cl, 
    felner@bgu.ac.il, 
    sven.koenig@uci.edu
}

\begin{document}

\newcommand\algname[1]{\textsf{#1}\xspace}
\newcommand{\ignore}[1]{}

\newcommand{\OS}[1]{{\textcolor{red}{\textbf{OS:} #1}}}
\newcommand{\todo}[1]{{\textcolor{magenta}{#1}}}
\newcommand{\newText}[1]{{\textcolor{red}{#1}}}
\newcommand{\good}{{\textcolor{blue}{\checkmark}}}

\newcommand{\myemph}[1]{{\color{teal}\emph{#1}}}

\maketitle

\begin{abstract}
Multi-objective search (MOS) has emerged as a unifying framework for planning and decision-making problems where multiple, often conflicting, criteria must be balanced. While the problem has been studied for decades, recent years have seen renewed interest in the topic across AI applications such as robotics, transportation, and operations research, reflecting the reality that real-world systems rarely optimize a single measure. 
This paper surveys developments in MOS while highlighting cross-disciplinary opportunities, and outlines open challenges that define the emerging frontier of MOS research.
\end{abstract}



\section{Introduction}
Multi-objective Search (MOS) problems are pervasive in real-world settings where decision makers must balance several, often conflicting, objectives. 
For example, in route finding applications, we are interested in simultaneously minimizing both travel time and fuel consumption, or distance and toll costs.
In many such cases, improvements in one objective cannot be achieved without hinderring another objective, making the search for well-balanced solutions both challenging and essential.

When a decision maker can articulate how much loss in one objective is acceptable for a given gain in another, all objectives can be turned into one scalar value by, e.g., optimising a weighted sum or another order-preserving (monotone) aggregation.
Then, the resulting problem can be solved by any standard single-objective algorithm. 
This aggregation approach, however, presupposes reliable \emph{apriori} information about acceptable trade-off for the decision maker, which is often not available to the algorithm.

An alternative approach to addressing the multidimensional trade-off is to use MOS  algorithms that compute the best attainable trade-offs wherein no objective can be improved without degrading at least one other objective. This set can then be presented to the decision maker for an a posteriori  preference articulation and final choice. 
%

%

While being a decades-old problem~\cite{vincke1976,hansen-1980,climaco2012multicriteria,current1993multiobjective,skriver2000classification,tarapata2007selected,ulungu1991multi}, in recent years, the study of MOS has attracted growing attention across multiple research communities. 
Dedicated workshops and tutorials addressing complex, often conflicting, objectives
have been featured in mainstream AI
venues 
(e.g.,
AAAI~[\citeyear{aaai24_tutorial}], 
IJCAI~[\citeyear{ijcai23_workshop,ijcai25_tutorial}],
AAMAS~[\citeyear{aaai24_tutorial}],
ICAPS~[\citeyear{icaps24_tutorial}],
ECAI~[\citeyear{ecai25_workshop}]
and 
SoCS~[\citeyear{socs23_tutorial}]),
and in robotics and machine-learning venues 
(e.g., RSS~[\citeyear{rss25_workshop}]
and
NeurIPS~[\citeyear{neurips24_workshop}]).
Related developments are also emerging in operations research (OR), transportation science, and evolutionary computation, where multi-objective optimization has a long tradition but is now being revisited with modern heuristic search, reinforcement learning, and hybrid approaches. This convergence of interests reflects a shared recognition that real-world decision-making rarely optimizes a single criterion, and that principled multi-objective reasoning is essential for building intelligent, robust, and adaptable systems.

\paragraph{Scope.}
In this paper, we highlight recent advances in the field in terms of problem variants, algorithms, applications and emerging directions. It is by no means a comprehensive literature review but an attempt to provide an accessible starting point for any researcher interested in  the field. 

Here, we focus on the setting of multi-objective \myemph{search}. However, we also highlight extensions and variants such as those that include  uncertainty.
Importantly, due to lack of space, we maintain a high-level description of approaches and refer the reader to~\cite{SalzmanF0ZCK23} for a technical overview of recent MOS advances.

\section{Problem Setting \& Variants}
\subsection{Notation}
Boldface font indicates vectors, lower-case and upper-case symbols indicate elements and sets, respectively. The notation~$p_i$ will be used to  denote the $i$'th component of~$\mathbf{p}$. 
The addition of two $d$-dimensional vectors $\mathbf{p}$ and~$\mathbf{q}$ and the multiplication
of a real-valued scalar $k$ and a $d$-dimensional vector $\mathbf{p}$ are defined  as $\mathbf{p}+\mathbf{q}=(p_1+q_1,\ldots, p_d+q_d)$ and
$k\mathbf{p}=(kp_1,\ldots,kp_d)$, respectively. 

Let $\mathbf{p}$ and $\mathbf{q}$ be $d$-dimensional vectors.
For a minimization problem, we say that~$\mathbf{p}$ 
\myemph{dominates}~$\mathbf{q}$ and denote this as~$\mathbf{p} \preceq \mathbf{q}$ if $\forall i, p_i \leq q_i$.
We say that $\mathbf{p}$ is \myemph{lexicographically  smaller} than~$\mathbf{q}$ and denote this as $\mathbf{p} \prec_{\rm{lex}} \mathbf{q}$ if~${p}_k < {q}_k$ for the first index $k$ s.t. ${p}_k \neq {q}_k$.
Finally, 
let~$\mathbf{p}$ and $\mathbf{q}$ be  two $d$-dimensional vectors
and let~$\boldsymbol{\varepsilon}$ be another $d$-dimensional vector such that $\forall i~\varepsilon_i \geq 0$.
We say that~$\mathbf{p}$ \myemph{approximately dominates}~$\mathbf{q}$ with an \myemph{approximation factor} $\boldsymbol{\varepsilon}$ and denote this as $\mathbf{p} \preceq_{\boldsymbol{\varepsilon}} \mathbf{q}$ if $\forall i, p_i \leq (1+\varepsilon_i) \cdot q_i$.

\subsection{MOS---Problem definition and variants}
\label{subsec:pdef}
\begin{figure}[t]
    \centering
    \includegraphics[width=0.275\textwidth]{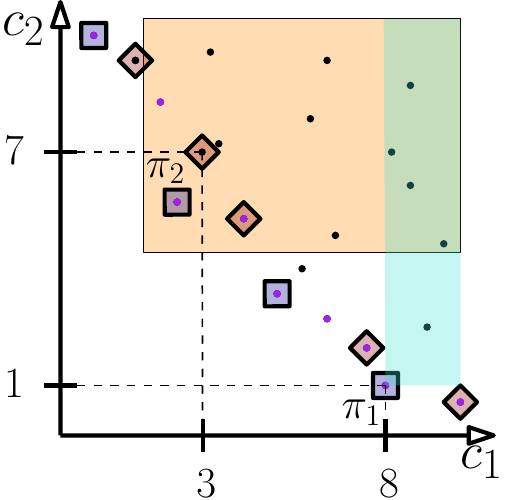}
\caption{Visualization of key MOS concepts for the special case of a bi-objective problem.
    Solutions on and not on the PF are visualized as purple and black dots, respectively.
    Visualization of all solutions dominated and approximately dominated by solutions $\pi_1$ and  $\pi_2$ are visualized by turquoise and orange regions respectively.
    Example for sets of solutions that approximate the PF which lie and which do not lie on the PF are depicted with purple squares and red diamonds, respective.
    Finally, the dominance factor $\textsc{Df}(\pi_1, \pi_2)$ in this example is
    $\max(\max(8/3-1,0),\max(1/7-1,0))=5/3$.
}
\label{fig:pf}
\end{figure}

In most variants of a MOS problem
a directed graph~$G=(V,E)$ 
where each edge $e\in E$ has a nonnegative cost vector $\mathbf{c}(e)\in\mathbb{R}^d$ where $d>0$ is the number of objectives. For the specific cases where $d=1$ and $d=2$, we refer to the problem as \emph{single-objective} and \emph{bi-objective}, respectively.
For a  path $\pi=\langle v_1, \ldots, v_k\rangle$ where $(v_i, v_{i+1}) \in E$, its cost $\mathbf{c}(p)$ is the sum of the edge costs. That is, $\mathbf{c}(\pi) = \sum_i \mathbf{c}(v_i, v_{i+1})$.
%
%
Given start and target vertices $s,t \in V$, 
a path from $s$ to $t$ is called a \emph{solution}. 
A solution is  \myemph{Pareto-optimal} iff its cost is not dominated by any other solution. 
See Fig.~\ref{fig:pf}.

\paragraph{Exact MOS.}
In the basic MOS problem, we are given vertices $s,t \in V$ and the goal is to compute the set $\Pi^\star$ of Pareto-optimal solutions, also known as the \myemph{Pareto front} (PF)~\cite{SalzmanF0ZCK23}.
Importantly, computing $\Pi^\star$ is \NP-hard~\cite{S87} as its cardinality may be exponential in~$|V|$~\cite{Ehrgott05,breugem2017analysis}. Even
determining whether a path belongs to~$\Pi^\star$ is \NP-hard~\cite{PY00}.

In certain settings, we would like to compute the PF from a source $s\in V$ to \emph{every} other vertex $v \in V$ (see, e.g.,~\cite{martins-1984,CasasSB21,KurbanovCV23})
or from \emph{any} vertex $u\in V$ to \emph{every} other vertex $v \in V$ 
(see, e.g.,~\cite{ZhangSFKUK23,CuchyVJ24}).

\paragraph{Approximate MOS.}
In real-world settings, we are often not interested in the entire PF as it may be too large to present to decision makers (for example, there may be thousands of solutions in the PF of large road networks~\cite{ren2025emoa}). 
Thus, 
we are often interested in computing a bounded approximation of $\Pi^\star$.
Here, we are given an approximation factor~$\mathbf{\varepsilon}$.
The \myemph{$\mathbf{\varepsilon}$-approximate PF} $\Pi^\star_{\mathbf{\varepsilon}}$ is a set of solutions such that 
$\forall \pi \in \Pi^\star, 
\exists \pi' \in \Pi^\star_{\mathbf{\varepsilon}}
~s.t.~
\mathbf{c}(\pi') \preceq_{\boldsymbol{\varepsilon}} \mathbf{c}(\pi)$.
Namely, every solution in $\Pi^\star$ is approximately dominated by some solution in  in~$\Pi^*_{\mathbf{\varepsilon}}$. 
Importantly, 
(i)~the $\mathbf{\varepsilon}$-approximate Pareto-optimal solution set is not necessarily unique
and  
(ii)~some variants of this definition require that $\Pi^\star_{\mathbf{\varepsilon}} \subseteq \Pi^\star$ while others don't (See Fig.~\ref{fig:pf}). 
%
Alternatively, some problem formulations seek 
a small representative set of solutions in~$\Pi^\star$~\cite{RiveraB022} (without any formal definition of ``small'').

\paragraph{Anytime MOS.}
Many applications benefit from obtaining a subset $\tilde\Pi^\star$ of $\Pi^\star$ as fast as possible. As more time is available to the algorithm, additional solutions from $\Pi^\star \setminus \tilde\Pi^\star$ are added to   $\tilde\Pi^\star$. The algorithm terminates when either (i)~the decision maker or the algorithm that uses the solutions terminates the algorithm or (ii) the entire PF has been returned (i.e., $\tilde\Pi^\star = \Pi^\star$). 
Formally, we define the \textit{dominance factor} of a solution $\pi$ over another solution $\pi'$ as 
$$
\textsc{Df}(\pi, \pi') = \max_{i=1,2\ldots N}\left( \max\left\{ \frac{c_i(\pi)}{c_i(\pi')} - 1\right\}, 0 \right),
$$
which measures how  ``good'' $\pi$ approximates $\pi'$. 
\textsc{Df}$(\pi, \pi')$ encodes the smallest $\varepsilon$-value that satisfies $\pi \preceq_{(\varepsilon,\ldots,\varepsilon)} \pi'$ (See Fig.~\ref{fig:pf}).
For a set of solutions $\Pi$, we define the \textit{approximation error} of $\Pi$ as 
$$\textsc{Err}(\Pi) = 
    \max_{\pi'\in \Pi^\star} 
        \{
            \min_{\pi \in \Pi} \textsc{Df}(\pi, \pi')
        \},
$$
which, roughly speaking,  measure the solution in $\Pi^\star$ that is ``least'' approximated by any solution in $\Pi$.

Now, to measure the performance of an anytime MOS algorithm, we typically wish to minimize the  \textit{Area Under the Curve} (AUC) of the approximation error formally defined as
$\textsc{Auc} = \int_{0}^{t^\text{limit}} \textsc{Err}(\Pi(t))$, where $t^\text{limit}$ is the runtime limit  and $(\Pi(t))$ is solution set returned at time $t$.

\paragraph{Incremental \& Dynamic MOS.}
When the MOS problem is applied in an online fashion (i.e., planning is interleaved with taking actions) and the query is updated (either because the target is updated or because the environment's representation is updated), one may want to avoid calling an algorithm from scratch and instead reuse previous search efforts.
Incremental multi-objective graph search algorithms  (see, e.g.,~\cite{RenRLC22} reuse previous  searches to speed up subsequent exact or approximate MOS searches.

Similarly, in MOS applications such as flight planning, dynamic traffic roadmaps, and telecommunication and data networks, the underlying graph changes over time since either its structure (edges, nodes) or the cost functions (weights, travel times, risks, etc.) evolve. 
In contrast to the incremental setting, here we are given the dynamics before planning begins and need to account for the temporal changes. This was only recently formulated by~\citet{CasasBKS21}. For additional details, see also the recent work by~\citet{ShovanKD25}.

\subsection{Beyond MOS}
\label{subsec:beyond}
While MOS assumes a deterministic model,  many real problems demand richer models. To capture stochasticity,  or general optimization beyond paths, several extensions  have been studied.
Each extends MOS along a different axis while keeping Pareto optimality central. We briefly describe each model, highlighting the similarities and differences compared to MOS.

\paragraph{Handling uncertainty.}
Recall that a MOS problem is defined using a graph $G = (V,E)$ together with edge costs~$\mathbf{c}$ which present a deterministic model. 
%
A \emph{Multi-objective Stochastic Shortest Path} (MOSSP) problem extends the MOS framework by introducing probabilistic transitions between states~\cite{RoijersWhiteson2017}. 
Formally, we are given a graph $G = (V, E)$  together with edge costs~$\mathbf{c}$ and a transition probability distribution over successor vertices. 
A policy $\mu$ maps vertices to successor edges, inducing a distribution over paths from the start $s$  to the target~$t$. 
The cost of a policy is defined as the expected cumulative cost vector across objectives. 
As in MOS, policies are compared using dominance: a policy $\mu_1$ dominates~$\mu_2$ if it has no worse expected cost in every objective and is strictly better in at least one. 
The goal is to compute a coverage PF of policies.

A Multi-objective Markov Decision Process (MOMDP) generalizes MOSSP by adopting an MDP formalism. In contrast to MOSSP which focuses on reaching a target in a stochastic graph with vector costs, MOMDP allow arbitrary horizon settings (e.g., finite, infinite, discounted) and sequential decision-making under uncertainty, not just reaching a target vertex.


\paragraph{Learning.}
\emph{Multi-objective Reinforcement Learning} (MORL) extends the MOS framework to settings where the agent interacts with an environment through repeated trial-and-error learning rather than having an explicit model of the state-transition dynamics~\cite{RLguide22,FTD24}. 
Formally, MORL is defined over the same structure as an MOMDP, $M = (S, A, P, \mathbf{c})$, with state space $S$, action space $A$, transition function $P$, and vector-valued cost or reward function $\mathbf{c}(s,a) \in \mathbb{R}^d$. 
However, unlike MOMDPs, in MORL the transition probabilities and reward distributions are not assumed to be known a priori. 
Instead, the agent learns a policy $\mu: S \to A$ through experience, typically by interacting with the environment and receiving vector-valued feedback.

MORL generalizes MOS in that it seeks Pareto-optimal solutions across multiple objectives, but unlike MOS, it does not assume a static graph with deterministic edges. 
Relative to MOSSP and MOMDP, MORL replaces planning with learning: instead of computing Pareto-optimal policies from a known model of uncertainty, the agent must discover them through exploration and approximation. 
Thus, MORL inherits the challenges of both reinforcement learning (e.g., exploration-exploitation trade-offs, function approximation) and MOS (e.g., dominance checks). 
The goal in MORL remains to approximate the Pareto set of policies, but learning algorithms must balance sample efficiency, preference sensitivity, and scalability in high-dimensional state and objective spaces.



\paragraph{Multi-objective Optimization.}
In this paper, we focus on MOS which can be seen as an instance  of the more general \emph{multi-objective optimization} (MOO) problem, (see, e.g.,~\cite{branke2008multiobjective,miettinen2012nonlinear,RoijersWhiteson2017,hwang2012multiple,emmerich2018tutorial}).
It is important to highlight the similarities and differences between the two fields.

MOO is the most general formulation of problems in which several, possibly conflicting, objectives must be optimized simultaneously. 
Formally, given a feasible set $\mathcal{X}$ and objective functions $f_i : \mathcal{X} \to \mathbb{R}$ for $i=1,\ldots,d$, the problem is to find the set of non-dominated solutions 
\[
\mathcal{X}^\star = \{x \in \mathcal{X} \mid \nexists y \in \mathcal{X},\ \mathbf{f}(y) \preceq \mathbf{f}(x),\ f(y) \neq f(x)\},
\]
where $\mathbf{f}(x) = (f_1(x),\ldots,f_d(x))$. 
Here, $\mathcal{X}^\star$ corresponds to the \myemph{Pareto set} whose image in $\mathbb{R}^d$ is the PF.

Relative to MOS, MOO generalizes the underlying domain: whereas MOS is defined over paths in a deterministic graph with additive vector costs, MOO is agnostic to the structure of the feasible set and can capture continuous, combinatorial, or black-box domains. 
Compared to stochastic settings such as MOSSP and MOMDP, MOO does not necessarily assume probabilistic dynamics or sequential decision processes; instead, it focuses purely on the optimization of static or offline-defined objectives. 
In contrast to MORL, MOO assumes direct access to the objective functions rather than learning them through interaction. 
In this sense, MOO serves as the broad umbrella under which MOS, MOSSP, MOMDP, and MORL can be seen as structured subclasses with additional constraints on the representation of $\mathcal{X}$, the dynamics of decision-making, and the information available to the algorithm.

From an algorithmic point of view, in contrast to MOS, which builds upon search algorithms,
MOO typically builds upon local and global optimization methods such as
genetic algorithms~\cite{deb2002fast,deb2013evolutionary,zhang2007moea},
particle swarm optimization~\cite{coello2002mopso},
and
simulated annealing~\cite{li2011adaptive}.

\ignore{
Evolutionary multi-objective optimisation (EMO) typically consider the optimization 

remains a strong option when objectives are numerous or black-box, and when gradient information is unreliable. Contemporary surveys emphasise three families: dominance-based (e.g., NSGA variants), indicator-based (e.g., hypervolume-driven), and decomposition-based (e.g., MOEA/D), with fresh attention to \emph{constrained} and \emph{dynamic} settings and to many-objective difficulties (selection pressure, loss of resolution, and visualisation) \cite{liang-tevc-2023-cmoea,sato-2023-maosurvey,wang-2023-emo-search-strategies,edmo-survey-2022}. In practice, EMO scales naturally across cores/GPUs and tolerates irregular evaluation costs; parallel MOEAs and island models are now mature \cite{falcon-cardona-2021-pmoea}. For MOS on graphs, EMO serves as (i) a front compressor/warm start for search (seed A* variants with a small, diverse set of trade-offs), (ii) an outer loop over scalarisations, and (iii) a robust baseline for $d\!>\!3$ where exact/anytime graph search becomes label-bound. Benchmarks from the EMO community (many-objective suites, constrained MOPs) are valuable for stress-testing search procedures beyond bi-criteria road networks \cite{weise-mostaghim-2022-benchmark}.

\paragraph{Classical planning}
\OS{does not really fit here}

In the domain of AI planning, most works only address
single-objective problems.
\citet{KhouadjiaSVDS13} suggest a domain-independent Multi-objective satisficing algorithm
which uses
an Evolutionary Algorithm (EA) to evolve sequences of partial
states for the problem at hand, calling an embedded algorithm
to solve in turn each subproblem of the sequence.

This comes in contrast to the lion's share of the literature about multi-criteria/objective
planning (see, e.g.,~\cite{GereviniSS08}) which optimize  an aggregation of the objectives.
}

\section{Algorithmic Advances}
In this section we outline recent algorithmic advances in MOS. We start with a brief historical overview and continue to outline tools, techniques and algorithms that advances the state-of-the-art in MOS. We conclude with a brief description of advances in generalizations of MOS (Sec.~\ref{subsec:beyond}) that have close ties to MOS algorithms

\subsection{Brief historical overview}
Early work on MOS established the algorithmic framework which is the basis of most modern  algorithms~\cite{hansen-1980,martins-1984,warburton-1987}. 
For efficiently computing~$\Pi^\star$, two notable approaches emerged.
The first generalizes the label-correcting paradigm to the multi-objective setting~\cite{guerriero2001label}. Label-correcting is an iterative shortest-path method that repeatedly updates tentative distance labels of vertices whenever a shorter path is found, allowing multiple updates per vertex until no further improvements are possible. 
The second generalizes  the celebrated \algname{A$^*$} algorithm~\cite{HNR68} which we detail next as most of the recent advancements fall under this category.

A notable contribution was the work by Stewart et al.~(\citeyear{stewart1991multiobjective}), who introduced 
Multi-Objective A* (\algname{MOA$^*$}).  
\algname{MOA$^*$} served as the foundation to multiple extensions (see, e.g.,~\cite{de2005new,mandow2010multiobjective}) which differ in 
which information is contained in the nodes, 
how nodes are ordered in the priority queue 
and 
how dominance checks are implemented and when they are performed (upon generation or upon expansion).
A key insight that dramatically improved the efficiency of these algorithms was to order the nodes in the priority queue in increasing lexicographic order and apply the notion of \myemph{dimensionality reduction}~\cite{pulido2015dimensionality}. See~\cite{SalzmanF0ZCK23} for an overview of the approach. 
The resulting algorithm based on this idea was termed \algname{NAMOA-dr}\footnote{Here, `dr' stands for dimensionality reduction.}.

%

Early approaches~\cite{warburton-1987,perny2008near,TZ09,breugem2017analysis} to approximating~$\Pi^\star$ focused on Fully Polynomial Time Approximation Schemes\footnote{An FPTAS  is an approximation scheme whose time complexity is polynomial in the input size and also polynomial in $1/\varepsilon$ where~$\varepsilon$ is the approximation factor.} (FPTAS)~\cite{V01}.
Unfortunately,
running these approaches on moderately-sized graphs (i.e., with roughly $10,000$ vertices) is often impractical~\cite{breugem2017analysis}.

\subsection{Algorithmic advances in MOS}

\paragraph{Exact approaches.}
In recent years, several algorithms dramatically improved the efficiency of exact MOS algorithms (see, e.g.,~\cite{CasasKSB23,CasasSB21,AhmadiTHK21,ren2025emoa}).
Notable examples that reduce the computational complexity of key operations include
(i)~\algname{BOA$^*$}~\cite{hernandez2023simple} which  adapted and simplified \algname{NAMOA-dr} for the bi-objective setting performing dominance checks in $O(1)$, 
and
(ii)~recent work~\cite{HSFKUK24,ren-arxiv22-emoa} which improves node indexing and data structures for dominance checks of two and three objectives to yield dramatic speedups.
Finally, recent work~\cite{AhmadiSHJ24} considered the more general set of graphs with negative edges.

\paragraph{Approximate approaches.}
\citet{GS21} suggested \algname{PPA$^*$}, an extension of \algname{BOA$^*$}  that introduced new pruning techniques to efficiently compute an {$\mathbf{\varepsilon}$-approximate PF} $\Pi^\star_{\mathbf{\varepsilon}}$ for the bi-objective setting.
\algname{PPA$^*$} was later generalized by~\citet{zhang2022pex} who suggested the \algname{A$^*$pex} algorithm which allows to compute $\Pi^\star_{\mathbf{\varepsilon}}$  for any number of objectives. The efficiency of \algname{A$^*$pex} stems from the observation that paths whose cost is very similar can be grouped in an efficient manner allowing to dramatically prune the PF.
\algname{A$^*$pex}  was later used to exploit correlation of edge costs~\cite{HFKS25},
develop anytime MOS algorithms~\cite{zhang-socs24-aapex} 
and
more~\cite{HSFKK24}, 
see also Sec.~\ref{sec:toolbox}.

\paragraph{Parallelization.}
While there has been some research on parallelizing  MOS algorithms (see, e.g.,~\cite{sanders2013parallel,erb2014parallel,medrano2015parallel}, this research direction has been largely unexplored~\cite{SalzmanF0ZCK23}.
%
%
%
Two notable exceptions include
(i)~The work by~\citet{ahmadi-2025-parallel-moa} who explored  permutations of objective orderings in parallel while sharing bounds to collapse subproblems, achieving near-linear speedups on many-core systems
and
(ii)~the work by~\citet{Ulloa0KFS24} who suggest an approach to compute set dominance checks or SDC (a key procedure, which dominates the running time of many state-of-the-art MOS algorithms) in parallel.
They exploit vectorized the operations offered by ``Single Instruction/Multiple Data'' (SIMD) instructions to perform SDC on ubiquitous consumer CPUs thereby dramatically improving the runtime of existing MOS algorithms.

\paragraph{Theory.}
There have not been many recent theoretical advances. 
\citet{CasasBKS21}  suggested an FPTAS for the new setting of dynamic MOS in which edges cost can change online. 
\citet{SkylerSAFSC0K0U24} extended theory from single-objective search (SOS) to MOS which characterizes the set of vertices and search nodes that any unidirectional search algorithm must expand to prove the optimality of the solution. 
Specifically, they introduce a classification of vertices into \emph{must-expand}, \emph{maybe-expand}, and \emph{never-expand} categories.
The notable difference between SOS and MOS is that vertices must be expanded to 
(i)~prove that any path in a PF is Pareto-optimal (these are called optimality vertices)
and 
(ii)~ensure that there are no more solution costs that are not represented in the PF (these are called completness vertices).
Completeness vertices  have no analogy in SOS.

\paragraph{Heuristics.}
Key to the success of heuristic search in general, and heuristic MOS  in particular, is the ability to incorporate domain knowledge using heuristics that guide the algorithm.
Almost all MOS algorithms  use the ``ideal point heuristic''~$\mathbf{h_{\rm ideal}}$, which combines a set of $d$ single-objective heuristics $h_1, \ldots, h_d$. 
Here, $h_i:V\rightarrow \mathbb{R}_{\geq 0}$ corresponds to the shortest path from each vertex according to the $i$'th objective  and $\forall v \in V~\mathbf{h_{\rm ideal}}(v) := ( h_1(v), \ldots, h_d(v) )$.

However, in contrast to  SOS, in the general case of MOS, the heuristic value of a vertex $v$ is not a single cost vector, but a \myemph{set} of \myemph{cost vectors} (see~\cite{mandow2010multiobjective} for original definition and~\cite{SalzmanF0ZCK23} for an in-depth discussion). 
While such heuristics, called Multi-Value Heuristics (MVH) are much more informative, the overhead of computing and using MVHs in MOS algorithms can be large and the total runtime is often larger than when using~$\mathbf{h_{\rm ideal}}$~\cite{geisser2022admissible}.

A notable example where MVHs are used is the recent work by~\citet{ZhangSFKSUK23}, which generalize Differential Heuristics (DHs)~\cite{goldberg2005computing}, a class of memory-based heuristics for SOS, to bi-objective search, resulting in Bi-Objective Differential Heuristics (\algname{BO-DHs}). They propose several techniques to reduce the memory usage and computational overhead of \algname{BO-DHs},  demonstrating  reductions in runtime of a bi-objective search algorithm by up to an order of magnitude.

\subsection{Algorithmic advances in MOS extensions}
While there has been many advances in MOSSP, MOMDP and MOO algorithms, which are not the focus of this paper, here we mention work that is closely related to MOS.
Recent work by~\citet{chen-aaai23-molao} suggests adapting  heuristic-search algorithms (which are the foundation of MOS algorithms) for MOSSP. This is done  by extending (single-objective) stochastic shortest-path algorithms, such as \algname{LAO*}~\cite{HansenZ01a} and \algname{LRTDP}~\cite{BonetG01}, to the multi-objective setting.
They also study how to guide their algorithms with domain-independent heuristics to account for the
probabilistic and multi-objective features of the problem.

\ignore{
\section{Algorithmic advances in MO extensions}

\paragraph{Handling uncertainty.}
Multi-objective stochastic shortest-path (MOSSP) problems generalize the stochastic shortest-path framework to optimise multiple objectives (such as time, fuel, and risk) simultaneously, while deferring preference aggregation because users cannot supply fixed weights a priori~\cite{RoijersWhiteson2017}.
Arguably, the most common variant of MOSSP are Multi-objective Markov Descision Problems (MOMDPs) which can be solved via inner-loop planning, which adapts Stochastic Shortest Path (SSP) solvers by generalising their operators to the multi-objective setting, or via outer-loop planning, which repeatedly solves scalarised versions of the problem using an SSP solver~\cite{RoijersWhiteson2017}.

Recent work by~\citet{chen-aaai23-molao} suggests adapting heuristic-search methds for MOSSP by extending foundational SSP algorithms such as \algname{LAO*}~\cite{HansenZ01a} and \algname{LRTDP}~\cite{BonetG01} to the multi-objective setting.
Moreover, they also study how to guide their newly-proposed algorithms with domain-independent heuristics differing by how they account for the
probabilistic and multi-objective features of the problem.

Finally~\citet{LiJS25} study the geometric structure of the PF in MOMDPs and report that the PF is on the boundary of a convex polytope whose vertices all correspond to deterministic policies, and neighboring vertices of the PF differ by only one state-action pair of the deterministic policy, almost surely. This allows to suggest a new algorithm which drastically reduces the complexity of searching for the exact PF.

\paragraph{Multi-Objective Reinforcement Learning (MORL)}
Recent research in Multi-Objective Reinforcement Learning (MORL) has moved beyond classic scalarized or value-vector approximations to more flexible and scalable methods. 
A notable advance is \emph{distributional Pareto-optimal MORL (DPMORL)}, which learns policies that respect not only expected returns but also the full return distributions---capturing uncertainty and risk preferences in settings such as autonomous driving~\cite{DPMORL}. 
Another emerging direction is the use of \emph{Lorenz dominance} to design fair and scalable MORL algorithms, ensuring equitable outcomes across objectives and demonstrating scalability in large transport domains such as Xi’an and Amsterdam~\cite{abs-2411-18195}. 
To address the curse of dimensionality in many-objective environments, reward-dimension reduction techniques dynamically transform the objective space while preserving Pareto-optimality, enabling learning with up to sixteen simultaneous objectives~\cite{ParkS25}.

Complementing these works, preference-aware  learning is gaining traction. 
For example,
\emph{Preference-Controllable RL (PCRL)} trains a single policy conditioned on user-provided preferences, enabling fine-grained control over which region of the PF is followed and improving both controllability and solution diversity~\cite{yang2025preference}.

Together, these advances reflect a shift toward methods that explicitly handle uncertainty, fairness, scalability, and user preferences in MORL. 
}

\section{MOS as an Algorithmic Toolbox}
\label{sec:toolbox}
Recently the algorithmic toolbox developed for MOS has also proven useful in other domains. 
Some, are new variants of MOS while others are seemingly unrelated optimization problems where  MOS approaches have been  useful.

\paragraph{Multi-objective minimum spanning tree.}
The Multi-Objective Minimum Spanning Tree (MO-MST) problem generalizes the classical MST problem to settings where edges are labeled with cost vectors. Instead of a single spanning tree with minimal total weight, the goal is to identify a Pareto set of spanning trees that represent undominated trade-offs among objectives. However, unlike  MST, for which there are polynomial time
algorithms that solve it,  MO-MST is NP-hard~\cite{fernandes2020empirical}.
MO-MST is important for  communication networks, where spanning trees must balance latency, bandwidth and resilience, and in transport and logistics, where constructing infrastructure with multiple cost criteria is essential (see, e.g.,~\cite{abs-1102-2524}).
MO-MST algorithms borrow heavily from MOS techniques (see, e.g.,~\cite{SourdS08, fernandes2020empirical,CasasSB25}).

\paragraph{MOS with objective aggregation.}
In many real-world problems with multiple objectives, the objectives interact in a complex manner, leading to problem formulations that do not allow out-of-the-box  usage of MOS algorithms~\cite{FuKSA23,SlutskyYWF21,provably_safe}.
Roughly speaking, this is because the search algorithms needs to treat differently paths that are and that are not solutions. 
For example, in robot inspection planning~\cite{FuKSA23,AlpertSKS25}, a robot is required to view as many points of interest (POI) as possible using an on-board sensor while minimizing path length.
The two objectives which define a solution $\pi$ are
the  number of POIs viewed along $\pi$ and the length of $\pi$.
However, every path that is not a solution must keep track of which POI was viewed, essentially defining a binary objective for each POI. This is because two paths to the same vertex that viewed different POIs cannot dominate one another as their final bi-objective cost depends on which POIs will be viewed in the future.

This creates a  mismatch between objectives at intermediate nodes, which we term \emph{hidden objectives}, and objectives at solution  nodes, which we term \emph{solution objectives}.
The relation between solution objectives and hidden objectives is captured via some method of \emph{objective aggregation}~\cite{PWAS25}. 
Returning to our inspection-planning example, 
there is one hidden objective that corresponds to each POI as well as one for path length
and there are two solution objectives corresponding to number of POIs viewed and path length.
Here objective aggregation is done by adding all (binary) cost values of POI hidden objectives.
We call such problems \emph{MOS with objective aggregation} (MOS-OA).
Importantly, MOS-OA algorithms can naturally employ the MOS algorithmic toolbox. Indeed, early versions of \algname{A$^*$pex}  were developed in the context of MOS-OA~\cite{FuKSA23}.

\paragraph{Multi-objective Multi-Agent Path Finding.}
The Multi-Agent Path Finding (MAPF) problem~\cite{SternSFK0WLA0KB19} involves finding non-colliding paths for multiple agents from their start locations to their respective target locations in a shared environment. The primary goal is to optimize a metric such as the sum of travel time of all agents or the makespan (i.e., task completion time). 
The Multi-Objective MAPF (MO-MAPF) problem extends the MAPF problem to multiple, often conflicting, optimization criteria such as makespan, energy consumption, safety margin, or fairness among agents. 
The result is not a single plan but a PF of MAPF plans, each representing a different trade-off. 
Recent algorithms (see, e.g.,~\cite{RenRC21a,RenRC21b,RenRC23,Wang0K024}) integrate MAPF and MOS to obtain scalable algorithms for this purpose.

\paragraph{Constrained shortest path.}
In the Constrained Shortest-Path problem (CSP)~\cite{storandt2012route} we are interested in computing a shortest path subject to some constraints (e.g., limited energy consumption for an autonomous agent).
This setting was generalized by \citet{skyler2022bounded} who consider the setting where we need to find a solution which belongs to~$\Pi^\star$  whose costs are below given upper bounds on each objective.
Later~\citet{HSFKK24} considered a similar setting but where we need to find a solution which belongs to~$\Pi^\star_{\mathbf{\varepsilon}}$ for some $\mathbf{\varepsilon}>0$.

\paragraph{$k$-Shortest simple path.}
In the $k$-Shortest Simple Path ($k$-SSP) problem, we are given a graph $G=(V,E)$ with regular (scalar) edge costs. Given start and target vertices $s,t \in V$ and a parameter $k$, we are tasked to compute the $k$  shortest paths between $s$ and $t$.
While this is a single-objective problem, recently~\citet{CasasSBH25} have shown that the $2$-SSP can be solved by a reduction to a bi-objective search problem.

\section{Emerging Applications}

We briefly review several diverse domains where MOS and its variants have been recently used. This showcases the  applicability of MOS  despite its relative simplicity when compared to the richer models reviewed in Sec.~\ref{subsec:beyond}.

\paragraph{Automated design \& synthesis.}
MOS has been applied to design problems in chemistry, biology, and engineering. 
One example is retrosynthesis planning in computational chemistry, which is the problem of finding reaction sequences that produce a target molecule.
\citet{LAI2025100130} consider multiple objectives and, by  searching for un-dominated synthesis routes, they were able to present several candidate pathways to a human chemist to evaluate and choose from.
Similarly, MOS found applications in drug discovery and generative design. For example, \citet{southiratn2025combimots} suggested a bi-objective search algorithm for generating molecular structures that balance affinity to two proteins while also satisfying drug-like property constraints The result is a set of novel
molecular candidates with high predicted efficacy and acceptable pharmacological profiles. which traditional single-objective or scalarized approaches would have likely missed.

\paragraph{Multi-modal journey planning.}
Multi-modal journey planning determines  routes combining different transport modes~\cite{BastDGMPSWW16}, which  inherently involves multi-objective optimization such as time, cost and  comfort.
These methods (see e.g.,~\cite{PotthoffS22a,PotthoffS22b}) build upon  MOS algorithms to make queries tractable at metropolitan scale.
Many real-world uses of such algorithms have  recently been documented. 
For example, OpenTripPlanner 2 is an open-source multi-modal journey planner for public transportation in combination with bicycling, walking, and mobility services such as  bike share and ride hailing. It has been deployed nationwide in Norway and Finland. In Portland (Oregon), it  provides about 40,000 trip plans on a typical weekday~\cite{otp-deployments}.

\paragraph{Robotics.}
In robotics, multiple objectives often need to be simultaneously balanced (e.g., cost, energy and safety)
In Sec.~\ref{sec:toolbox} we discussed  robot inspection planning in the context of MOS-OA. 
Another  example is autonomous vehicle (AV) planning using rulebooks~\cite{SlutskyYWF21,CensiSWYPFF19,HA25,penlington2024optimizationrulebooksasymptoticallyrepresenting},  where the system must generate a trajectory that complies with a set of potentially conflicting traffic rules. Consider, for instance, Singapore’s Final Theory of Driving that requires 
(i) maintaining at least a one-meter gap when passing a parked vehicle and (ii)~prohibits crossing a solid double white lane divider. When an AV encounters a car improperly parked along such a divider, it may be impossible to satisfy both requirements simultaneously. 
Fortunately, requirements often form in a hierarchy---e.g., avoiding a collision is more important than keeping safety margin from parked vehicles and than maintaining lane.
Rulebooks are a systematic way to address such settings.
Here, a rule corresponds to  an objective and a rulebook defines a hierarchy that induces a partial order. For example rule $r_1$ (avoiding collision) is more important than rules $r_2,r_3$ (maintaining safety distance and lane) but rules $r_2,r_3$ are incomparable.
This  generalizes MOS which is a ``flat'' hierarchy where no objective (rule) is more critical than any other one.


\ignore{
\paragraph{Learning user preferences}
A recent and exciting application that we are currently exploring is using MOS to learn user preferences.
Consider a surgeon performing trajectory planning for needle biopsy (we are considering the specific setting of neurological surgery). Through years of experience the expert surgeon  knows what she cares about---distance from blood vessels, from anatomical obstacles, trajectory length, entry angle and more. But she can't articulate how she accounts for each factor. We would like to \emph{learn} her specific cost function (protocols differ between countries, hospitals and even surgeons) by suggesting trajectories and asking for a ranking (i.e., which path is preferred). We then use this ranking as labeled data to learn the surgeons latent cost function.
We (i) can only offer a small number of trajectories at any instance to account for the cognitive load of comparing trajectories and (ii) can't ask for many examples as the surgeon's time is extremely precious.
What we do know is that solutions that we should offer lie on the PF of what the surgeon cares about. This focuses the process only to trajectories that may be useful.
}

\section{Open Challenges and Opportunities}

Despite the  progress reviewed in this paper, several fundamental challenges remain open.
In contrast to~\citet{SalzmanF0ZCK23}  who discuss technical challenges that are the foundations for advancing  MOS algorithms, here we focus on  challenges and opportunities that will increase the \emph{impact} of MOS.

\paragraph{Scalability and dimensionality.}
Most existing algorithms scale poorly when the number of objectives grows beyond two or three. Approximate and bounded suboptimal MOS algorithms  partially address this issue, but there is no consensus on how to effectively navigate high-dimensional cost spaces which may be essential in real-world applications.

\paragraph{Dynamic and uncertain environments.}
Real-world deployment increasingly requires algorithms that adapt to changing graphs or stochastic models. While recent works study dynamic MOS, MOSSP and MOMDP, current algorithms mostly remain  theoretical or are limited to small instance sizes. Developing practical, general-purpose dynamic MOS algorithms is an interesting research opportunity.

\paragraph{Preference elicitation and user modeling.}
In many applications, decision makers cannot provide trade-offs upfront. Integrating preference elicitation into the search process—by interactively presenting Pareto-optimal candidates and learning from user choices—remains an underexplored yet impactful research direction. Combining MOS algorithms with methods from preference learning and human-in-the-loop AI is another  research opportunity.

\paragraph{Cross-fertilization between research communities.}
Important opportunities exist at the interface of MOS and other AI subfields. In reinforcement learning, MORL is rediscovering many algorithmic ideas from MOS; conversely, MOS can benefit from policy-gradient and distributional methods. 
In many domains such as robotics, large-scale transport systems and OR, multiple objectives are prevalent but existing MOS formulations need to be adapted to be applied effectively.

\ignore{
\begin{itemize}
    \item measure diversity
    \item exploit correlation
    \item mvh
    \item {Scaling to many objectives.} Beyond $d{\approx}3$, fronts and path sets blow up. We need principled compression (reference-point and $\varepsilon$-nets; path sharing) with fast dominance and clear decision relevance.
\end{itemize}
}

\paragraph{Benchmarks.}
Classical MOS benchmarks focus on road networks and grid worlds, while robotics emphasizes motion-planning roadmaps, and reinforcement learning relies on synthetic MO-MDPs. This fragmentation hampers comparison across research communities.

A standardized benchmark suite that spans classical MOS, stochastic and dynamic settings, and application-inspired domains (such as transportation, robotics and chemistry) would be a major step forward. Beyond static datasets, benchmarks should include interactive tasks for preference elicitation and evaluation metrics that reflect both efficiency and effectiveness. The community would benefit from a shared repository of graphs, environments, and evaluation protocols to foster reproducibility and comparability.

\ignore{
\begin{itemize}
    \item 1https://download.geofabrik.de/europe/germany/bayern.html
	\item https://download.geofabrik.de/europe/germany.html
	\item http://users.diag.uniroma1.it/challenge9/download.shtml
	\item GRID + EXP + HFPP (FPTAS paper)
    \item many-objective suites \cite{weise-mostaghim-2022-benchmark}
\end{itemize}
}

\section{Conclusion}
MOS has rapidly expanded from a niche research topic to a broad principle that influences many disciplines and applications and studied by multiple communities. On the algorithmic side, there have been  significant improvements in exact search, new approximate and parallel algorithms, and theoretical insights.
On the applications side, numerous communities have started to formulate their problems in terms of trade-offs between different metrics  and adopt MOS
methods to handle these trade-offs. 
While MOS was the focus of this paper, there should be more cross fertilization between different multi-objective optimization approaches.

\section*{Acknowledgments}
This research was supported by Grant No. 2021643 from the United States-Israel Binational Science Foundation (BSF).

\bibliography{aaai2026}

\end{document}